\documentclass[10pt,journal,compsoc]{IEEEtran}
\ifCLASSOPTIONcompsoc
  \usepackage[nocompress]{cite}
\else
  \usepackage{cite}
\fi
\usepackage{graphicx}				
\ifCLASSINFOpdf
\else
\fi
\usepackage{amsmath}
\usepackage{amssymb}
\usepackage{textcomp}

\begin{document}
%
\title{Dynamic User Segmentation and Usage Profiling}
%
%
%
%

\author{Animesh~Mitra,
        Saswata~Sahoo,
        and~Soumyabrata~Dey,
\IEEEcompsocitemizethanks{\IEEEcompsocthanksitem A. Mitra is a Research Assistant at the University of Florida, USA.\protect\\
E-mail: animesh18mitra@gmail.com
\IEEEcompsocthanksitem S. Sahoo is an Associate Director at Flipkart, India. \protect\\
E-mail: ssaswata@yahoo.co.in
\IEEEcompsocthanksitem S. Dey is an Assistant Professor at Clarkson University, USA. \protect\\
E-mail: sdey@clarkson.edu
}
\thanks{Manuscript received December 20, 2017.}}

%
%

\markboth{Journal of \LaTeX\ Class Files,~Vol.~14, No.~8, August~2015}%
{Shell \MakeLowercase{\textit{et al.}}: Bare Demo of IEEEtran.cls for Computer Society Journals}
%



\IEEEtitleabstractindextext{%
\begin{abstract}
Usage data of a group of users distributed across a number of categories, such as songs, movies, webpages, links, regular
household products, mobile apps, games, etc. can be ultra-high dimensional and massive in size. More often this kind of data is categorical and sparse in nature making it even more difficult to interpret any underlying hidden patterns such as clusters of users. However, if this information can be estimated accurately, it will have huge impacts in different business areas such as user recommendations for apps, songs, movies, and other similar products, health analytics using electronic health record (EHR) data, and driver profiling for insurance premium estimation or fleet management.

In this work, we propose a clustering strategy of such categorical big data, utilizing the hidden sparsity of the dataset. Most traditional clustering methods fail to give proper clusters for such data and end up giving one big cluster with small clusters around it irrespective of the true structure of the data clusters. We propose a feature transformation, which maps the binary-valued usage vector to a lower dimensional continuous feature space in terms of groups of usage categories, termed as covariate classes. The lower dimensional feature representations in terms of covariate classes can be used for clustering. We implemented the proposed strategy and applied it to a large sized very high-dimensional song playlist dataset for the performance validation. The results are impressive as we achieved similar-sized user clusters with minimal between-cluster overlap in the feature space (8\%) on average). As the proposed strategy has a very generic framework, it can be utilized as the analytic engine of many of the above-mentioned business use cases allowing an intelligent and dynamic personal recommendation system or a support system for smart business decision-making.
\end{abstract}

\begin{IEEEkeywords}
Clustering, High Dimensional Data, User Segmentation, Sparsity, Usage Profiling.
\end{IEEEkeywords}}

\maketitle

\IEEEdisplaynontitleabstractindextext

%
\IEEEpeerreviewmaketitle

\IEEEraisesectionheading{\section{Introduction}\label{sec:introduction}}

%
%
%
%
\IEEEPARstart{M}{ost} business intelligence tasks involve the collection of usage data to mine the behavioral pattern of the customers. It helps to devise key business strategies which are profitable for customers and business owners at the same time. Improved business strategies lead to better revenue models and outcomes. On the other hand, insights into customers' usage and behavioral pattern lead to enhanced user experience in terms of personalized recommendations and services. In this work, we are particularly interested in data on the population of users distributed across a number of usage categories. The usage categories can be potentially any tangible product, application, and service such as songs played, movies watched, mobile apps installed, products
added to an online shopping cart, traded financial stocks, clicked weblinks, etc. Such datasets are invariably massive in size and a potential number of usage categories can be very large, and hence, the usage data vectors are very high dimensional. Each entry of the observation vector on the usage assumes one of the two distinct values 0 or 1 depending on the absence or presence of a category. In this work, we focus on clustering this type of datasets to find groups of homogeneous users with the predominant presence of certain common usage categories in each cluster. We propose a novel method for efficient clustering of high dimensional data set with sparse and categorical (binary-valued) variable. Most relevant works in this context fail to cluster such usage data meaningfully and in most cases, give one very large cluster and small clusters around it irrespective of the true structure of the data clusters. This is
specifically because none of the clustering algorithms to date is capable of handling all three stated characteristics of the
data. However, some of the methods are worth mentioning as they can handle some of the mentioned data features. For example, sparse SVD \cite{ref1_Aharon2006} can handle sparse and large-size data sets, whereas ROCK\cite{ref2_Guha1999} (A Robust Clustering Algorithm for Categorical Attributes) is good for large-sized and categorical data, CLIQUE \cite{ref3_Zaki1997} and CURE \cite{ref4_Guha1998} are efficient for large sized and high dimensional data. Furthermore, DBSCAN(A Density-Based Algorithm for Discovering Clusters)\cite{ref5_Ester1996} is good for sparse data, K-modes \cite{ref6_Huang2003} and WARD \cite{ref7_Ward1963} is effective for categorical data, PAM(Partitioning around medoids) \cite{ref8_Park2009}, BIRCH(Balanced Iterative Reducing and Clustering using Hierarchies) \cite{ref9_Zhang1996}, EM(Expectation Maximization) \cite{ref10_Dempster1977} for large-sized data and DENCLUE(Density-based clustering) \cite{ref11_Hinneburg2007}, Chameleon \cite{ref12_Karypis1999} for high dimensional data, iterative k-means \cite{ref19_rao2018, ref20_Rao2021} on big data. In support of our statement, we provide a comparison of the performances of our algorithm and some of the existing methods. The results clearly show the failure of other algorithms for our problem. Following is the list of advantages of our algorithm over the existing methods.
\begin{enumerate}
	\item To the best of our knowledge, this is the first work for handling large-sized, high-dimensional data sets with sparse categorical variables.
	\item Our algorithm has a computational advantage as we can train on a subsample of the data and apply the trained model on full-size data.
	\item Our method can easily assign a new user to its appropriate cluster in runtime.
\end{enumerate}

Recommendation: Once a user is mapped to a cluster, cluster-specific information can be used for generating user profiles which in turn can be utilized for enhanced personalized recommendations and services. The profiles consist of a statistical prediction model on future usage patterns, customer propensity models over multiple product categories, etc. These profiles play a critical role in recommendations on products and services, particularly in connection to app usage, online movie, and music services, e-commerce, etc.

The rest of the paper is organized as follows. In section 2 we described the problem framework and characteristics of our data. In section 3 we described the detailed methodology. Section 4 is used for showing the application of our method on a real data set (Million songs). Sections 5 and 6 are respectively the performance comparison with other methods and discussion.

\vspace{1.5cm}
\IEEEraisesectionheading{\section{Framework}\label{sec:framework}}
Consider $p$ categories $A_1, A_2, ..., A_p$ and an associated $p$ dimensional random vector, denoted by $X^{p\times1} = (X_1, X_2, ..., X_p)^{'}$, the $j$th component $X_j$ assuming two distinct values, $1$ or $0$ according as a presence or absence of the $j$th category $A_j$ . Binary responses on the presence or absence of the $p$ categories are available as observations on the random vector $X$ on $n$ subjects $U_1, U_2, ..., U_n$. Observed value of the random vector $X$ for the $i$th subject $U_i$ is denoted by $\tilde{x}_i = (x_{i1}, x_{i2}, ..., x_{ip})^{'}$. The complete data set is essentially given by $\{\tilde{x}_i = {(x_{ij})^p}_{j=1}, i=1, 2, ..., n\}$ and can be represented by an $n\times p$ dimensional observation matrix denoted by $D = (\tilde{x}_1, \tilde{x}_2, ..., \tilde{x}_n)^{'}$.

We define a few notations and operations that we shall be referring to in due course. An indicator variable $\mathcal{I}$ is defined as a logical function, such that $\mathcal{I}(u \in A) = 1$ or $0$ according to $u \in A$ or not. We define a distance metric $d(., .) : {{\{ 0, 1\} }^p}\times{{\{ 0, 1\} }^p}$ $\,\to\,$ $[0, 1]$ where for two $p$ dimensional ${\{ 0, 1\} }^p$ valued vectors $(x = x_1, x_2, ..., x_p)^{'}$ and $(y = y_1, y_2, ..., y_p)^{'}$, $d(x, y) = p^{-1} \sum\limits_{k=1}^p \mathcal{I}(x_k \neq y_k)$. For an $N$ dimensional vector $V$ with elements from real
line, $V = (V_1, V_2, ..., V_N)$, the $L_2$ vector norm is defined as $\parallel V\parallel {}_2 = (\sum\limits_{i=1}^N {V_i}^2)^{(1/2)}$ and for a rectangular matrix $M = (M_{ij})$ with $a$ rows and $b$ columns, the $L_{\infty}$ matrix norm is given as $\parallel M\parallel {}_{\infty} = Ma{x^a}_{i=1} \sum\limits_{j=1}^b \mid M_{ij} \mid$. The cardinality of a set $S$ is an integer-valued set function $\nu(.)$ and is denoted by $\nu(S)$.

\subsection{High Dimensional Sparse Structure}\label{subsec:highDimensionalSparseStructure}
Referring to the observation matrix $\mathcal{D}$, note that $\sum\limits_{j=1}^p x_{ij} \leq p$ for all $i$. We are particular interested in a sparse and high dimensional $\mathcal{D}$, in the sense that  $\parallel \mathcal{D}\parallel {}_\infty = o(min\{n,p\})$, $\parallel {\mathcal{D}}^{'}\parallel {}_\infty = o(min\{n\,p\})$ and $n, p$ $\,\to\,$  $\infty $. Each row and
column of the observation matrix $\mathcal{D}$ is populated with lot many $0$’s and only very few $1$’s and hence the row sums and column sums of $\mathcal{D}$ are essentially very small in comparison to the matrix dimension. 

\subsection{Utilizing the Sparsity}\label{subsec:utilizingTheSparsity}
Here we try to discover important sources of variations among the observed values of $X$ in the binary-valued lattice structured feature space $\{0, 1\}^p$ of categories. We define a group of categories as a class. There might be a number of classes, say $K$ of them each comprising of a number of categories that explain the separation among the observations more prominently than the $p$ categories themselves (with possibly, $K << p$). Each of these classes combines a small amount of category-level variations and represents one important source of variation. In this work, we see whether this class-level variation can be discovered by pruning important category-level variations which are similar for decent subsets of observations. If a decent enough subset of observations is available such that they have similar categories responsible for creating separation from others then the subset of observations essentially indicates important class-level variations in the dataset. We shall discover subsets of this nature that jointly can summarize a number of important class-level variations and try to express original $0-1$ sparse observations in terms of these classes. The rest of the clustering will follow in the transformed feature space.

\vspace{1.5cm}
\IEEEraisesectionheading{\section{Methodology}\label{sec:methodology}}
In the language of feature transformation, for $p$ dimensional $X$, our goal is to look out for a transformation $\Phi(.) = (\phi_1(.), \phi_2(.)), ..., \phi_K(.))$ such that $\Phi : \{0, 1\}^p$ $\,\to\,$ $[0, 1]^K$. It transforms the $n \times p$ dimensional data matrix $\mathcal{D} = (\tilde{x}_1, \tilde{x}_2, ..., \tilde{x}_n)^{'}$ to a reduced $n \times k$ dimensional feature
matrix $\mathcal{D}_\Phi = (\Phi(\tilde{x}_1), \Phi(\tilde{x}_2), ..., \Phi(\tilde{x}_n))^{'}$. We introduce the concept of classes which are instrumental to develop the transformation $\Phi$.

\subsection{Classes}\label{subsec:classes}
We define a class as a subset of all possible categories $A_1, A_2, ..., A_p$. $A$ complete set of classes $\mathcal{C} = \{C_1, C_2, ..., C_K\}$ each explaining an independent and principal source of variation of $X$. The classes enable us to represent each $\tilde{x}_i$ in a reduced dimensional feature space by means of the transformation $\Phi$ with $\Phi(\tilde{x}_i) = (\phi_1(\tilde{x}_i), \phi_2(\tilde{x}_i), ..., \phi_K(\tilde{x}_i))$ where

\begin{equation} 
	\phi_u(\tilde{x}_i) = \frac{\nu(\{j:x_{ij}=1\} \cap \{j:A_j\in C_u\})}{\nu(C_u)}
	\label{eqn1}
\end{equation}

Note that $\Phi(.)$ transforms each $0-1$ string of length $p$ observed by the random vector $X$ to a weight vector of length $K$, giving the loading to different classes.

Each class is a group of categories and a principal variation component. First, we explore the sparse structure of variation at the actual observation space of the $p$ categories which we refer to as the category-level variation. Then category-level variations are suitably combined to find the classes. To discover the sparse structure in the category-level variations of the data we work with a sample of rows of $\mathcal{D}$ of suitable size leading to a sampled data matrix $\mathcal{D}^{*}$, say. If sampled properly, this is expected to cover a reasonable amount of the total variation of the dataset. One strategy could be to ensure $\mathcal{D}^{*}$ column averages are proportional to the respective column averages of $\mathcal{D}$ so that the sparse structure of $\mathcal{D}$ is retained in that of $\mathcal{D}^{*}$ up to a proportionality constant. We shall discuss an implementable sampling strategy later. For the time being, let us assume that we have a representative sample of rows of $\mathcal{D}$, given by $\mathcal{D}^{*}$, which has, $m$ rows and $p$ columns.

\subsection{Category Level Variations}\label{subsec:categoryLevelVariations}
The distances between any two pairs of observations have a sparse representation in terms of the category-wise dissimilarity. Utilizing this sparsity we investigate the categories that contribute more to the pairwise distances. In this connection, we refer to the sampled data matrix $\mathcal{D}^{*}$ and first try to find out the sparse structure of the variation at the $p$ dimensional observation space of categories.

The variations among the observed $\tilde{x}_i$’s have a sparse representation in terms of the components of $\tilde{x}_i$ in the sense that, the distance of each $\tilde{x}_i \in \mathcal{D}^{*}$, contrasted with the complementary part $\mathcal{D}^{*} - \{\tilde{x}_i\}$ is largely decided by only a few components of $\tilde{x}_i$. The rest of the components only add to the noise.  Associated to the $m\times p$ dimensional sampled data matrix $\mathcal{D}^{*}$, define a matrix $W = (w_{ij})$ of same dimension, such that $w_{ij} = 1$ or $0$ according as the $j$th component of $\tilde{x}_i$ is important explanatory feature for the distances $d(\tilde{x}_i, \tilde{x}_j), \tilde{x}_j \in \mathcal{D}^{*}, j\neq i$. We do the same investigation one by one for each $\tilde{x}_i \in \mathcal{D}^{*}$ with reference to the respective complementary part $\mathcal{D}^{*} - \{\tilde{x}_i\}$, i.e. run the same analysis for $i = 1, 2, ..., m$. We try to explain pairwise distances as a linear combination of component-wise dissimilarity of corresponding pairs of observations of $X$ as

\begin{equation} 
	\begin{split}
 		h(d(\tilde{x}_i, \tilde{x}_j)) = \sum\limits_{k=1}^p \beta_{ik} \mathcal{I}(x_{ik} \neq x_{jk}) + e_{ij}, \\
 		i = 1,2, ...,m, j=1,2, ...,m, i\neq j
	\label{eqn2}
	\end{split}
\end{equation}

Here $h(.)$ is a suitable link function that linearizes the distance between pairs of vector-valued observations as a function of dissimilarity between individual components of pairs of vectors. For each $i$, the coefficients $\tilde{\beta}_i = (\beta_{i1}, \beta_{i2}, ..., \beta_{ip})^{'}$ indicate the importance of the different categories towards explaining the distance of $\tilde{x}_i$ from the complementary sampled group of observations $\mathcal{D}^{*} - \{\tilde{x}_i\}$. We approximate the pairwise distances in terms of pairwise dissimilarity of only the important categories. The coefficients $\tilde{\beta}_i$ are estimated as
$\hat{\tilde{x}} $

\begin{equation} 
	\begin{split}
		\hat{\tilde{\beta}}_i = ArgMin_{\tilde{\beta}_i}\sum\limits_{j=1}^m (h(d(\tilde{x}_i, \tilde{x}_j)) - \sum\limits_{k=1}^p \beta_{ik} \mathcal{I}(x_{ik} \neq x_{jk}) )^2  \\
		+ \lambda \sum\limits_{k=1}^p \mid \beta_ik \mid 
	\label{eqn3}
	\end{split}
\end{equation}

with suitably chosen tuning parameter $\lambda > 0$ \cite{ref13_Geer2008}. Because of the $L_1$ penalty, the components of $\tilde{x}_i$ which do not contribute to deciding its distances from the complementary sampled group $\mathcal{D}^{*} - \{\tilde{x}_i\}$ would have $\beta_{ik}$$'$s shrank to 0. Running the optimization in $(3)$ for $i = 1, 2, ..., m,$ we estimate the $W$ matrix by $\hat{W}$ with ${\hat{w}}_{ij} = \mathcal{I}({\hat{\beta}}_{ij} \neq 0)$

\subsection{Class Level Variations}\label{subsec:classLevelVariations}

For the $m$ sampled rows in $\mathcal{D}^*$, the category-level variation components are listed in $W$ matrix. The $1'$s in the $i$th row of the $W$ matrix essentially indicate important categories $A_{j}'$s which contribute to the separation of $\tilde{x}_i$ with the rest of the $\tilde{x}_{i'} \in \mathcal{D}^*$. This is true for all $i = 1, 2, ...,m$. Here we find out subsets of observations such that, observations in each subset essentially have similar important categories approximating pairwise distances from other observations. A subset of observations of reasonable size with similar important categories essentially forms a group of categories to give a class-level variation component. With reference to
matrix $W$, overall, if there are $K$ class-level sources of variations, $W$ matrix should have nearly a $K$ rank structure. We describe, from the estimated matrix $\hat{W}$, an efficient method to discover the low rank structure of $\hat{W}$ and hence recover the classes $\mathcal{C} =\{C_1, C_2, ..., C_K \}$.

We define the following subsets of observations which will be important for the evaluation of the classes $\mathcal{C} =\{C_1, C_2, ..., C_K \}$. Hypothetically, consider a prospective class $C_u$, which will be having certain categories $\{A_{u1}, A_{u2}, ... \}$, say. The categories are of course unknown at this stage, and the goal essentially is to find them. Consider the corresponding subset of observations.

\begin{equation} 
\begin{split}
D_u = \{ \hat{\tilde{w}}_i \in \mathcal{D}^{*}: \hat{w}_{ij} = 1 \text{ for all j such that } A_j \in C_u \} \\
u = 1, 2, ..., K,   \hat{\tilde{w}}_i = (\hat{w}_{i1}, \hat{w}_{i2}, ..., \hat{w}_{ip})^{'}.
\label{eqn4}
\end{split}
\end{equation}

Our end goal is to evaluate ${C_u}'$s in such a way that each combines a number of category-level variations and represents a principal and independent source of class-level variation. Hence the categories that add to a class-level variation should be as far as possible distinctive from that of the other classes. The subset of observations $D_u$ carries category-level variations for categories included in the class $C_u$. To ensure maximal separation among the classes ${C_u}'$s, we evaluate $C_1, C_2, ..., C_K$ such that there is a maximum between class separation among the subsets $D_1, D_2, ..., D_K$. 

Moreover, we also need to fix the number of classes $K$. The classes should be so constructed that each of them has a reasonable uniform size and any two of them have minimal overlapping of categories that they include. The size uniformity ensures that each class has reasonable dimension reduction capacity in terms of absorbing a number of categories inside and at the same time, none of them absorbs too many dimensions either. The minimal overlapping ensures that each class is as far as possible independent and exclusive in terms of explaining the variations in the categorical feature space. To ensure the two, we choose the following two factors (1) Entropy difference and (2) Impurity. For $K$ classes, the entropy difference is defined as $M(K) = \sum\limits_{j=1}^K \pi_j log(\pi_j) - log(1/K) $ where $\pi_j$ is the proportion of categories out of the $p$ categories included in the $j$th class. A small value of this difference indicates more uniformity in the size of the classes. We also
define impurity of the classes $\mathcal{C} =\{C_1, C_2, ..., C_K \}$ as $I(K) = \frac{2}{K(K-1)} \sum\limits_{i=1}^K \sum\limits_{j=i+1}^K \frac{\nu(C_i \cap C_j)}{\nu(C_i \cup C_j}$. Clearly, less impurity in the classes shows better separation. Thus minimizing entropy difference and impurity is also our objective along with maximizing separations among $D_1, D_2, ..., D_K$.

With reference to the $K$ subsets defined in (4) the between-class variation of the rows of $\hat{W}$ matrix is given by

$ 
	\mathcal{B}_K(\mathcal{C}, \hat{W}) = \sum\limits_{u=1}^K \parallel \frac{\sum_{i:\hat{\tilde{w}}_i \in D_u}\hat{\tilde{w}}_i}{\nu(D_u)} - \frac{\sum_{u=1}^{K}\sum_{i:\hat{\tilde{w}}_i \in D_u}\hat{\tilde{w}}_i}{\sum_{u=1}^{K}\nu(D_u)} \parallel_{2}^{2}	
$
and the within class variation of the rows of $\hat{W}$ matrix is given by

$	
	\mathcal{W}_K(\mathcal{C}, \hat{W}) = \sum\limits_{u=1}^K \sum\limits_{i:\hat{\tilde{w}}_i \in D_u} \parallel \hat{\tilde{w}}_i - \frac{\sum_{i:\hat{\tilde{w}}_i \in D_u}\hat{\tilde{w}}_i}{\nu(D_u)} \parallel_{2}^{2}.
$

Tuning the upper bounds $\epsilon_1$ and $\epsilon_2$ for the impurity and entropy difference we carry out the following optimization to evaluate optimal $C_1, C_2, ..., C_K$:

\begin{equation} 
	\begin{split}
		Max_{K, C_1, C_2, ..., C_K} = [\mathcal{B}_K(\mathcal{C}, \hat{W}) - \mathcal{W}_K(\mathcal{C}, \hat{W})] \\
		\text{such that } I(K) < \epsilon_1, M(K) <  \epsilon_2.
	\label{eqn5}
	\end{split}
\end{equation}

Based on the optimized choice of the covariate classes $C_1, C_2, ..., C_K$, we carry out the feature transformation as described in (1) for all the observations corresponding to the rows of $\mathcal{D}$ to get a reduced $n\times K$ dimensional feature matrix $\mathcal{D}_\Phi = (\Phi(\tilde{x}_1), \Phi(\tilde{x}_2), ..., \Phi(\tilde{x}_n))^{'}$, where $ \Phi(\tilde{x}_i) = ( \phi_1(\tilde{x}_i), \phi_2(\tilde{x}_i), ..., \phi_K(\tilde{x}_i))^{'}$. These reduced dimensional feature values can be clustered using the K-Means algorithm. For $L$ clusters $E_1, E_2, ..., E_L$ of the $n$ observations, with respective cluster centroids being $\mu_1, \mu_2, ..., \mu_L$, the following optimization will directly follow

$ Min_{\mu_j, E_j, j=1,2, ...,L} \sum\limits_{j=1}^L \sum\limits_{\Phi(\tilde{x}) \in E_j} \parallel \Phi(\tilde{x}) - \mu_j \parallel_{2}^{2}$

\subsection{Sampling Strategy for $\mathcal{D}^*$}\label{subsec:samplingStrategy}
The core of the proposed feature transformation rests in the evaluation of the category-level variations components by a sparse linear representation of the distance in terms of individual category dissimilarity as given in (2). The category-level variation components are evaluated by shrinking the
category-wise dissimilarity components using LASSO for a training sample data matrix $\mathcal{D}^*$. Category-level variation components will lead to class-level variation components and these classes will eventually lead to lower dimensional feature transformation $\Phi$. The class level variation components evaluated based on the sampled data matrix $\mathcal{D}^*$ will be extendable to the complete data matrix $\mathcal{D}$ if the sparse structure of the whole data matrix $\mathcal{D}$ is reasonably retained in $\mathcal{D}^*$.Thus we propose a sampling strategy to choose $\mathcal{D}^*$ in such a way that the sparse structure of the data remains the same and that is achieved by keeping the column average and row average same as $\mathcal{D}$.

\subsection{Recommendation model}\label{subsec:recommendationModel}
Once the user segmentation is performed, user cluster-specific information can be utilized for further personalized recommendations and services. The easiest way of such recommendations can be a suggestion of any category associated with any other member of the same user cluster. But this method has an inherent shortcoming. First, the union of clusters associated even within a cluster is quite high. Recommending any such category might be less effective for a user. For the personalized recommendation of the $i$th user, we utilize two different pieces of information such as category-wise frequency
distribution within the cluster and the distance of other cluster members from the $i$th user. We combine these two pieces of information in a probabilistic framework to estimate the likelihood of recommending any category for $i$th user.

Suppose $C$ is a user cluster containing $l$ users $U_1, U_2, ..., U_l$. In aggregate, they use $A_1, A_2, ..., A_c$ categories. Our task is to provide recommendations for the $U_i (i = 1, ..., l)$. We construct the distance vector $\tilde{d_i} = (d_{i1}, d_{i2}, ..., d_{il})$ where $d_{ij} = dist(U_i, U_j)$ is euclidean distance between $U_i$ and $U_j$ in the reduced dimensional space. Since we prefer to use those users' usages which are closer to the $U_i$, we change the distance vector $\tilde{d}_i$ to similarity vector $\tilde{s}_i$ where $\tilde{s}_{ij} = \frac{\sum d_{ij} - d_{ij}}{\sum d_{ij}} \forall j = 1, 2, ..., l$. As a result of the transformation $\tilde{s}_i$ becomes a probability vector.

We have a frequency distribution of the categories $f_1, f_2, ..., f_c$. With proper scaling we change it to probability vector $\tilde{p}_i \text{ } \forall i = 1, 2, ..., c$. Thus we have now prior probabilities of the categories. Finally, we will give a score to each category given the target user is $U_i$. Optimum categories for the recommendation to the target user will be selected with the highness of the score.

The score of category $j$ for recommending user $i$ is given by $L(A_j\mid U_i)$. We know that $P(A_j\mid \bigcup\limits_{k=1\neq i}^{l} U_k) \propto P(\bigcup\limits_{k=1\neq i}^{l} U_k \mid A_j)P(A_j)$ and also  $P(\bigcup\limits_{k=1\neq i}^{l} U_k \mid A_j) = \sum\limits_{k=1\neq i}^{l} P(U_k \mid A_j) P(A_j)$

$P(A_j) = p_j$ and $P(U_k \mid A_j) \propto S(U_k \mid A_j)$ where $S(U_k \mid A_j) = s_{ik}$ if $U_k$ uses $A_j$ and $0$ otherwise.

Final score will be  $L(A_j \mid U_i) = \sum\limits_{k=1\neq i}^{l} S(U_k \mid A_j)P(A_j)$

So using the scores $L(A_j \mid U_i) \forall j$ we sort them and preferably higher scores will be taken for the recommendation.

\vspace{1.5cm}
\IEEEraisesectionheading{\section{Application on the real data}\label{sec:application}}
In this section, we investigate the performance of the proposed method of clustering in a real dataset on songs being played by a group of users. We take a subset of the 'Million song dataset' \cite{ref14_Mahieux2011} for the performance investigation. Here the list of users is given by $\{U_i, i=1, 2, ..., p \}$. In our investigation, the total number of unique users is $n = 1,00,000$ and the total number of unique songs is $p = 55,793$. We arrange the data set in an $n\times p$ dimensional matrix $\mathcal{D}$ with $(i, j)$th entry being $x_{ij}$, assuming $1$ or $0$ according to user $U_i$ has listened to the song $A_j$ or not. We fix $m = 3000$ to form the training sample data matrix $\mathcal{D}^*$ to estimate the classes. Optimizing the objective function in (3) we evaluate the $\tilde{W}$ matrix of dimension $m\times p$. Next, we move to class selection. 


\begin{table}[t] 
\begin{center}
\caption{Entropy change, impurity, Similarity Matrix $L_{\infty}$ with changes in number of clusters.} \label{table:Table_1}
\begin{tabular}{|c|c|c|c|}
  \hline
  No of     & Impurity  & Entropy       & Similarity Matrix\\
  cluster   &           & Difference    &\\
  (L)       & I(L)      & (M(L))        &   $\parallel S_L\parallel_{\infty}$ \\

  \hline  
   50       &   0.3846  &  0.3934   & 0.4440 \\
   100      &   0.2916  &  0.1475   & 0.3502 \\
   150      &   0.2330  &  0.0893   & 0.2901 \\
   200      &   0.1958  &  0.0650   & 0.2542 \\
   250      &   0.1656  &  0.0477   & 0.2243 \\
   300      &   0.1488  &  0.0847   & 0.2063 \\
   350      &   0.1321  &  0.0660   & 0.1871 \\
   400      &   0.1267  &  0.0721   & 0.1798 \\
   450      &   0.1153  &  0.0706   & 0.1671 \\
   500      &   0.1050  &  0.0759   & 0.1566 \\
   550      &   0.1053  &  0.0822   & 0.1533 \\
   600      &   0.0880  &  0.0851   & 0.1382 \\
   650      &   0.0918  &  0.0864   & 0.1420 \\
   700      &   0.0817  &  0.0812   & 0.1279 \\
   750      &   0.0779  &  0.0849   & 0.1231 \\
   800      &   0.0758  &  0.1011   & 0.1223 \\
   850      &   0.0718  &  0.0954   & 0.1152 \\
   900      &   0.0704  &  0.0834   & 0.1155 \\
   950      &   0.0690  &  0.0917   & 0.1131 \\
   1000     &   0.0646  &  0.1114   & 0.1085 \\
   1050     &   0.0610  &  0.1089   & 0.1031 \\
   1100     &   0.0599  &  0.1184   & 0.1019 \\
   1150     &   0.0586  &  0.1185   & 0.0986 \\
   1200     &   0.0585  &  0.1164   & 0.0969 \\
   1250     &   0.0551  &  0.1222   & 0.0962 \\
   1300     &   0.0529  &  0.1317   & 0.0923 \\
   1350     &   0.0508  &  0.1290   & 0.0904 \\
   1400     &   0.0489  &  0.1434   & 0.0879 \\
   1450     &   0.0485  &  0.1333   & 0.0863 \\
   1500     &   0.0472  &  0.1499   & 0.0848 \\
   1550     &   0.0462  &  0.1399   & 0.0843 \\
   1600     &   0.0454  &  0.1466   & 0.0837 \\
   1650     &   0.0449  &  0.1430   & 0.0813 \\
   1700     &   0.0413  &  0.1624   & 0.0789 \\
   1750     &   0.0417  &  0.1503   & 0.0778 \\
   1800     &   0.0411  &  0.1731   & 0.0784 \\
   1850     &   0.0399  &  0.1691   & 0.0755 \\
   1900     &   0.0397  &  0.1581   & 0.0748 \\
   1950     &   0.0382  &  0.1673   & 0.0739 \\
   2000     &   0.0379  &  0.1785   & 0.0733 \\  
   \hline
   
\end{tabular}
\end{center}
\end{table}

We need to fix the number of classes we will be settling into. The classes should be so constructed such that each of them has a reasonable uniform size and any two of them have minimal overlapping of songs. The minimal overlapping ensures that each class is as far as possible independent in terms of explaining the variations in the transformed feature space. To ensure the two, we choose the following two factors (1) Entropy difference and (2) Impurity as described in section 3.3. Clearly, the lesser both $M(K)$ and $I(K)$ are, the better it is for class selection, i.e., we prefer $K_1$ over $K_2$ if $M(K_1) < M(K_2)$ and prefer $K_3$ over $K_4$ if $I(K_3) < I(K_4)$. From Figure 1 we see that $I(K)$ has a downward trend although $M(K)$ has an upward trend with increasing $K$. So we need to fix thresholds for both functions of $K$ such that the first $K$ which satisfies both threshold conditions will be the number of classes to go with. We empirically fix: (1) $I(K) \leq 0.015$ (2) $M(K) \leq 0.30$. The first $K$ which satisfies both (1) and (2) is $175$. Thus using $K = 175$ we find our optimal classes as described in (5) in the previous section. Based on the selected classes we carry out the feature transformation as given in (1) and the K-Means clustering of the transformed dataset immediately follows. 

\begin{figure}[t]
	\centering
	\includegraphics[width=1\linewidth]{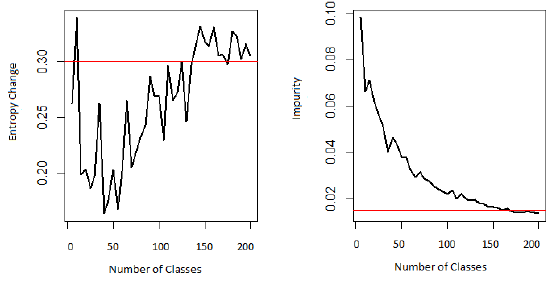}
	\caption{Choice of number of category classes ($K$) based on entropy difference ($M(K)$) and impurity function ($I(K)$), the thresholds being shown by red lines.}
	\label{fig:fig1}	
\end{figure}

The metrics which we use to determine the optimal number of clusters are the same as we had while selecting optimal classes of categories. For $L$ clusters given by $E_1, E_2, ..., E_L$, we consider the following two metrics:

\textbf{Entropy Change:} $M(L) = \sum\limits_{j=1}^{L} q_jlog(q_j) - log(1/L)$,  where $q_j$ is proportion of observations included in the $j$th cluster $E_j$.

\textbf{Impurity:} $I(L) = \frac{2}{L(L-1)} \sum\limits_{i=1}^{L} \sum\limits_{j=i+1}^{L} \frac{\nu(\tilde{E}_i \cap \tilde{E}_j)}{\nu(\tilde{E}_i \cup \tilde{E}_j)}$, where $\tilde{E}_i$ is the set of songs present in the $i$th cluster $E_i$. Note that, a small value of the entropy change $M(L)$ indicates that each cluster includes a reasonable number of observations and that the clusters are comparable in size. The smallest value $0$ of the entropy change indicates perfectly uniform-sized clusters. On the other hand, a small impurity $I(L)$ indicates that the clusters are less overlapped in terms of the songs being played by the users of each of the clusters. The aim essentially is to produce clusters of comparable sizes with very little overlap between any two pairs of clusters in terms of the songs being predominantly present in each of them. Hence, for good clustering, the two metrics should produce small values. However, one can expect an increasing trend in entropy change with an increasing number of clusters whereas for impurity, the trend will be opposite.

From Figure 2 we can easily see that the two functions are opposite in nature. So the point of intersection of the two curves could be a good solution to decide on the number of clusters. We choose $L = 750$ as the optimum number of clusters to go for.

Next, we investigate the performance of the clustering for a number of clusters, $L = 750$. In Figure 3, we display the number of observations out of $n = 1,00,000$ included in the $750$ clusters. We consider the cluster indices in the horizontal axis and the number of observations included in the different clusters are displayed by columns of proportional heights in the vertical axis. The sum of all the vertical column heights will be essentially $n = 1,00,000$. The clusters clearly turn out to be of comparable sizes and very close to uniformity. The entropy change value is as low as $0.08$ which also stands as good evidence in support of the near uniform cluster size distribution along with average impurity of $8\%$.

\begin{figure}[t]
	\centering
	\includegraphics[width=1\linewidth]{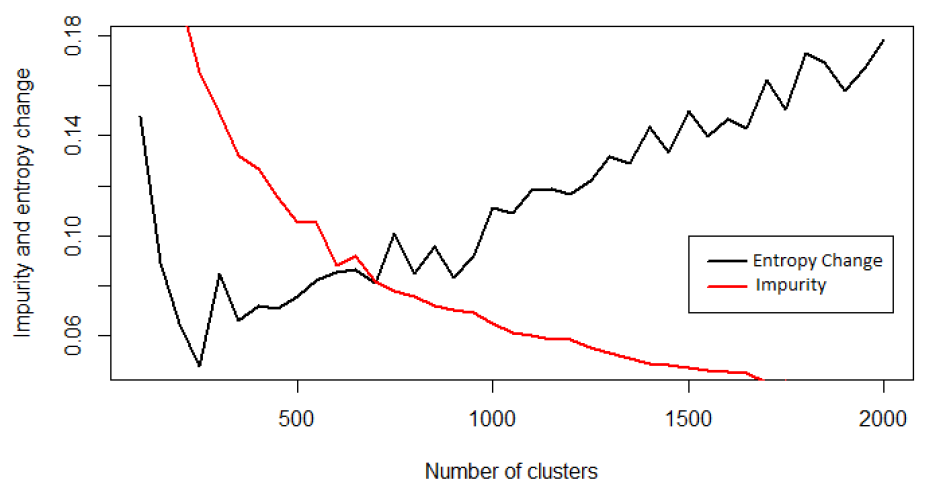}
	\caption{Selection of the Optimal number of clusters, on the basis of entropy change and impurity function. The optimal choice is $L = 750$ here.}
	\label{fig:fig2}	
\end{figure}

\begin{figure}
	\centering
	\includegraphics[width=1\linewidth]{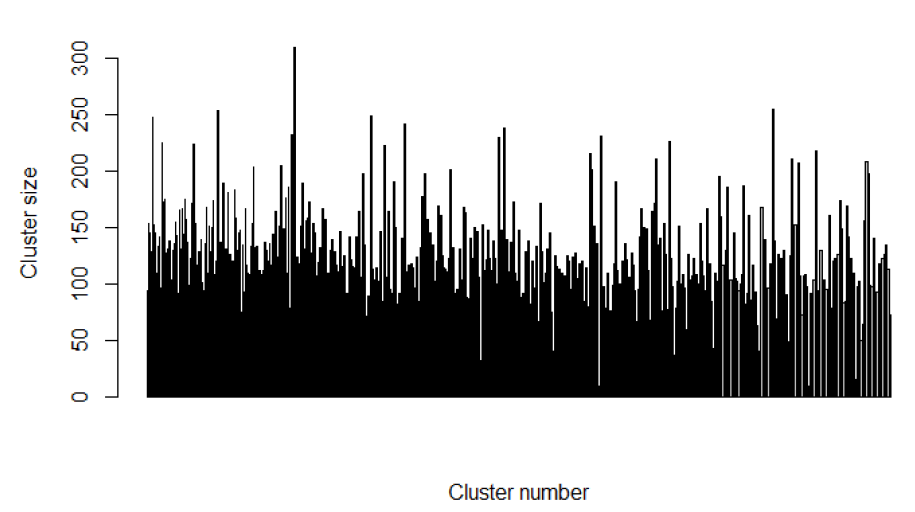}
	\caption{Cluster size distribution for $L = 750$ clusters where the cluster indices are in the horizontal axis and the number of observations included in the different clusters is displayed by columns of proportional heights in the vertical axis.}
	\label{fig:fig3}	
\end{figure}

Finally, we investigate the exclusivity of the clusters in terms of the uniqueness of songs being predominantly present in different clusters. Let us denote the number of songs that occur simultaneously in exactly $f$ of the clusters out of the $L$ clusters by $p(f)$. For a good clustering in terms of how exclusive and unique each of the clusters are depending on the songs each of them contains, we should expect that most songs should not be present in many clusters simultaneously and $p(f)$ should be a fast decaying function of $f$. In Figure 4 we confirm that actually this is the case. That means, any of the songs simultaneously appearing in many clusters is a very low probability event. We have seen that, out of all the $p = 55,793$ songs, $50\%$ songs have appeared in not more than $22$ clusters, $70\%$ songs have appeared in not more than $44$ clusters, and $90\%$ songs have appeared in not more than $108$ clusters at the same time.

We also investigate the quality of the clusters when the number of clusters, $L$ is varied within a range of $50$ to $2000$ with a difference of $50$. For $L$ clusters we consider the $L \times L$ dimensional cluster similarity matrix as $S_L =(S_{L}^{ij})$ with $S_{L}^{ij} = \frac{\nu(\tilde{E}_i \cap \tilde{E}_j)}{\nu(\tilde{E}_i \cup \tilde{E}_j)}$, where, as defined before, $\tilde{E}_i$ is the set of songs present in the $i$th cluster $E_i$. We replace the diagonal elements with $0$. Note that, the average of the $i$th row of this matrix indicates the average similarity of the $i$th cluster with the rest of the $L - 1$ clusters. Maximum of the row averages of these similarity matrices, which are essentially respective $L_{\infty}$ matrix norms $\parallel S_L\parallel_{\infty}$ are also indicators of cluster similarity. A small value of entropy changes $M(L)$, impurity $I(L)$, and $\parallel S_L\parallel_{\infty}$ are indicative of good clustering. varying $L$, we list the values of the three metrics in Table 1. Clearly, the larger the number of clusters, the better the cluster separation but more pure clusters come at the cost of less uniformity in the cluster size distribution. These observations are evident in Table 1. Hence, at the cost of losing some uniformity in the cluster sizes, it is possible to get more pure clusters with higher degrees of separations, i.e. impurity can be even less than $4\%$ with around $2000$ clusters. Nonetheless, we have been able to cluster the song data into a number of meaningful and well-separated clusters using the proposed methodology. There is a trade-off between the cluster size uniformity and overlapping and hence suitable selection of a number of clusters is crucial to control the two properties. The trade-off might depend on the specific use case one might be interested in. However, the dataset is of very high dimension and has extreme sparsity as we have considered in the present work. The performance evaluation results discussed so far do bring up a story of the successful implementation of the proposed feature transformation followed by clustering to a commendable extent.

\begin{figure}
	\centering
	\includegraphics[width=1\linewidth]{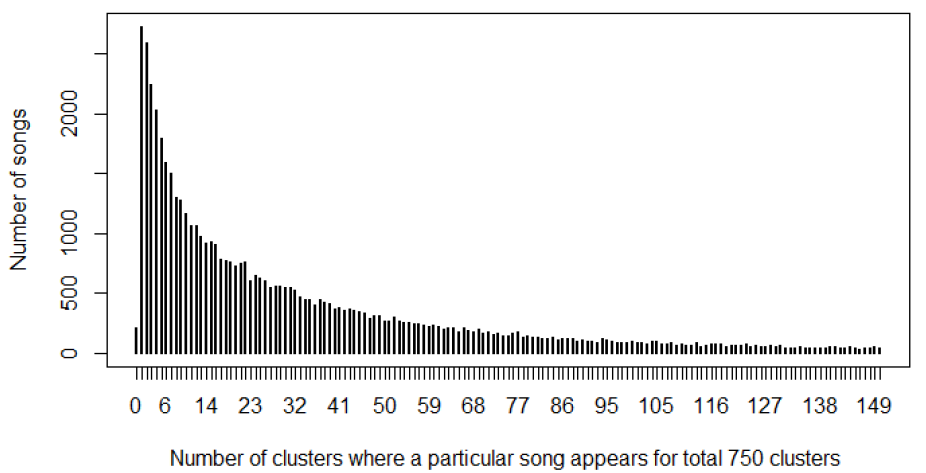}
	\caption{Column diagram for Number of songs, i.e $p(f)$ present only in $f$ clusters, for $f = 1, 2, ..., 150$.}
	\label{fig:fig4}	
\end{figure}

\vspace{1.5cm}
\IEEEraisesectionheading{\section{Comparison with the standard literature}\label{sec:comparisonWithLiteratire}}
In this section, we focus on comparing our algorithm with standard literature. First, we compare our algorithm with the popular clustering algorithms. We also compared our performance with the k-means clustering algorithm applied on the data projected on a lower dimensional space using popular dimension reduction techniques. Due to space and time complexity constraints, a subset of the Million song dataset is taken into account for all our comparisons. Among $1,00,000$ users $3,000$ users are sampled randomly keeping the length of each user vector unchanged and all the validations are done on this sampled data.

\subsection{Comparison with standard clustering techniques}\label{subsec:comparisonClustering}
Among the various clustering algorithm available in the literature we selected four of them for the purpose of performance comparison based on their popularity in the academic world and their relevance to our problem framework. The following are the mentioned algorithms: (1) K-means (2) Hierarchical-Ward (3) ROCK and (4) PAM (k-medoids). 

Among the four stated algorithms, k-means is the most commonly used. The method can handle very large datasets and is best suited for continuous data. Whereas, Ward is a hierarchical clustering algorithm that is robust for both categorical and continuous data but generally applicable for small-sized data. On the other hand, ROCK is especially applicable for categorical attributes, and can handle large datasets but strength is not in the cases of high dimensional sparse data sets. Finally, PAM is a k-means-like partitioning algorithm that is efficient for high dimensional data but not efficient for large and sparse data sets. All of the four mentioned algorithms have their specific areas of strength but none of them is particularly applicable for high dimensional, categorical, and sparse data sets. The experimental results reiterate the statement as all of the methods have failed to provide any meaningful clusters. As can be seen in Figure 5, each of the algorithms is converged with one big cluster accompanied by several small clusters resulting in a very non-uniformed distribution of cluster sizes. Looking at the clusters, it can be safely stated that non of the algorithm has performed well on our data set.

\begin{figure}[t]
	\centering
	\includegraphics[width=1\linewidth]{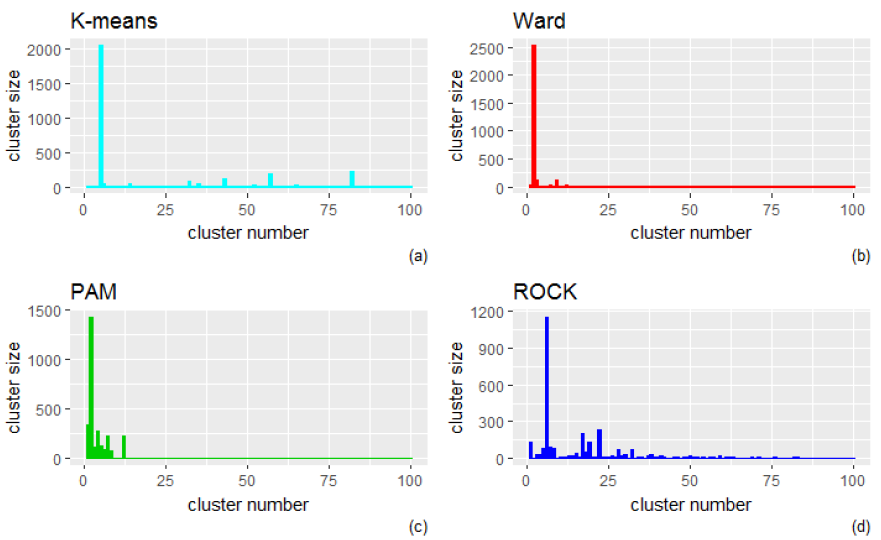}
	\caption{Performance of clustering using different techniques.}
	\label{fig:fig5}	
\end{figure}

\begin{figure}[t]
	\centering
	\includegraphics[width=1\linewidth]{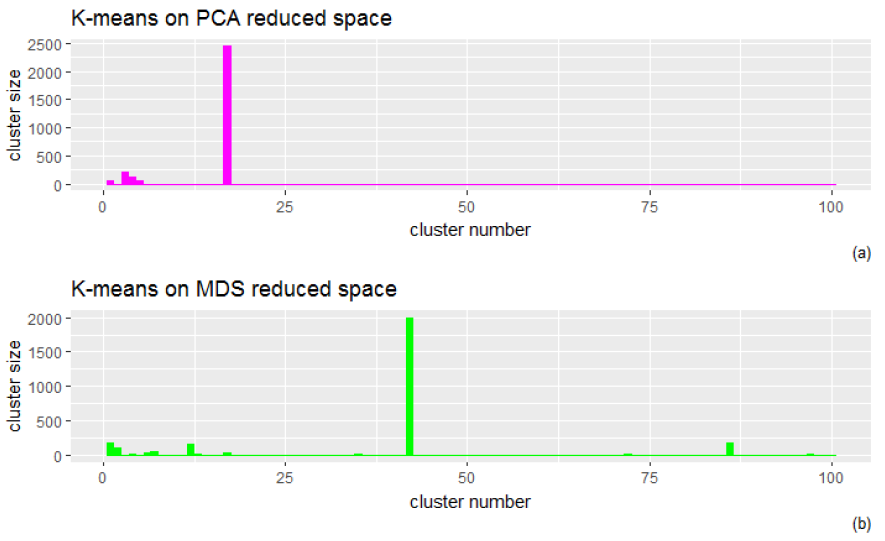}
	\caption{K-means Clustering after reducing the dimensions using standard techniques.}
	\label{fig:fig6}	
\end{figure}

\subsection{Performance of dimensionality reduction}\label{subsec:performanceDimensionalityReduction}
Due to the sparse structure of the large high dimensional observation matrix $\mathcal{D}$ it possibly has a very low-rank representation. However, discovering the low-rank structure of $\mathcal{D}$ is extremely difficult in terms of its eigen representation due to computationally infeasibility and mathematical inaccuracy. The sample size and the relative size of the eigenvalues have a positive influence on the accuracy of eigenanalysis whereas, an increase in dimension $p$ shows a negative influence on the same. Asymptotic properties of sample eigen values and eigen vectors depend on the joint influences of these positive and negative factors. In the absence of sufficient separation of the eigenvalues, very large $p$ may result in completely indistinguishable eigenvalue estimates and strongly inconsistent sample eigenvectors \cite{ref15_Nadler2008, ref16_Shen2013, ref17_Fan2015}. This may put principal component analysis(PCA) and Multidimensional Scaling (MDS) \cite{ref18_KruskalWish1978} under serious threat as both methods require eigenanalysis of a high dimensional matrix. PCA performs eigen decomposition on variance-covariance matrix of $\mathcal{D}$ (size $p \times p$), whereas MDS needs it
on distance matrix of $\mathcal{D}$ (size $n \times n$). Moreover, even if the eigen-decomposition can be performed accurately by some magical method, the eigenvectors' directions will be uninterpretable for the binary-natured observation space. This is because, for the categorical variables, the observation spaced is formed by a $\{0, 1\}^p$ lattice space where any points other than the lattice corners are undefined. Our claim of failure of dimension reduction techniques is demonstrated in Fig. 6. Both PCA and MDS are used to preserve $90\%$ of data variance which resulted in the dimension reduction from $55793$ categories to $1000$ and $500$ dimensions respectively. It can be seen that the k-means clustering on the reduced dimensional data gives very poor clusters (Fig. 6 (a), (b)). It can be safely concluded that any other clustering algorithm would have given a similar output.

\subsection{Performance of our proposed method}\label{subsec:performanceProposedMethod}
We first reduce the dimension of the data using the covariate class method and our optimal reduced dimension is $750$. We apply K-means to the concerned data and find that the clusters are more uniform in nature. Impurity is also very less which is shown in the previous section. From figure number 11 it can be seen it gives better clusters than the standard techniques which are already mentioned.

\begin{figure}[t]
	\centering
	\includegraphics[width=1\linewidth]{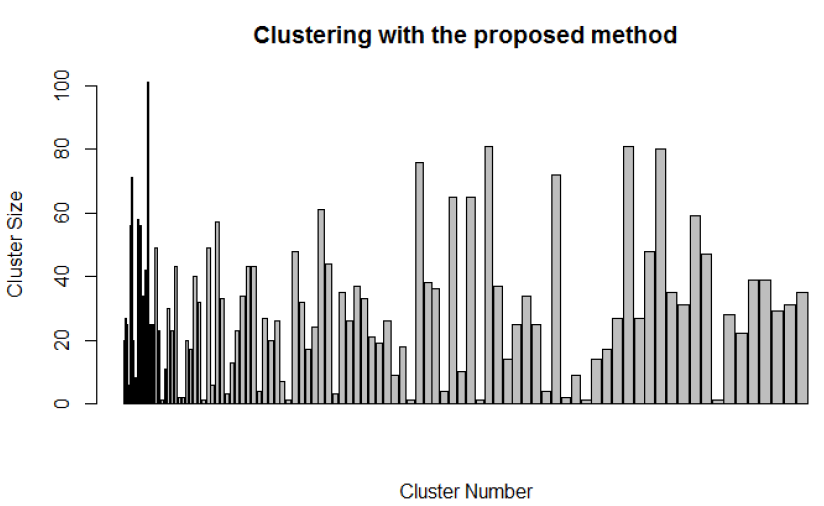}
	\caption{Clustering using the proposed method.}
	\label{fig:fig7}	
\end{figure}

\vspace{5.9cm}
\IEEEraisesectionheading{\section{Discussion}\label{sec:discussion}}
The data sets of the nature that we have discussed in this work, where the possible number of categories are too many but only a few of them are present for a given user is very common in a number of applications. In such a high dimensional sparse setup, we have tried to find out a sparse representation of the pairwise distance in terms of category-level dissimilarity. Subsets of observations having similar category-level variations are used to find important class-level variation components. These components are supposed to explain the majority of the variations present in the dataset. We map the sparse high dimensional observations to the lower dimensional space of these classes and this transformed feature map led us to commendable clustering of the data. As discussed, our algorithm has huge commercial potential in different application areas such as product recommendation in e-commerce, personalized recommendation systems for apps, movies, songs, optimized business decision making etc.

The application of the method on the real-world Million song dataset generated size-wise evenly distributed clusters with minimal overlap between class categories. We also suggested a recommendation procedure utilizing cluster-specific user profile information. But, due to the unavailability of ground truth in our dataset, we could not perform the validation of the recommendation system. We want to continue our work in this line to perform the validation on a dataset (such as an app usage dataset) with super-category level ground truth. This will give us the visibility of an end-to-end recommendation system. 

\ifCLASSOPTIONcaptionsoff
  \newpage
\fi



%




{\small
\bibliographystyle{ieee}
\bibliography{bibliogrpahy}

\begin{thebibliography}{10}\itemsep=-1pt

\bibitem{ref1_Aharon2006}
M.~Aharon, M.~Elad, and A.~Bruckstein.
\newblock K-svd: An algorithm for designing overcomplete dictionaries for
  sparse representation.
\newblock {\em IEEE Transactions on Signal Processing}, 54(11):4311--4322,
  2006.

\bibitem{ref14_Mahieux2011}
T.~Bertin-Mahieux, D.~P.~W. Ellis, B.~Whitman, and P.~Lamere.
\newblock The million song dataset.
\newblock In A.~Klapuri and C.~Leider, editors, {\em ISMIR}, pages 591--596.
  University of Miami, 2011.

\bibitem{ref10_Dempster1977}
A.~P. Dempster, N.~M. Laird, and D.~B. Rubin.
\newblock Maximum likelihood from incomplete data via the {EM} algorithm.
\newblock {\em Journal of the Royal Statistical Society: Series B}, 39:1--38,
  1977.

\bibitem{ref5_Ester1996}
M.~Ester, H.-P. Kriegel, J.~Sander, and X.~Xu.
\newblock A density-based algorithm for discovering clusters in large spatial
  databases with noise.
\newblock In {\em Proceedings of the Second International Conference on
  Knowledge Discovery and Data Mining}, KDD'96, page 226–231. AAAI Press,
  1996.

\bibitem{ref17_Fan2015}
J.~Fan and W.~Wang.
\newblock Asymptotics of empirical eigen-structure for ultra-high dimensional
  spiked covariance model, 2015.

\bibitem{ref4_Guha1998}
S.~Guha, R.~Rastogi, and K.~Shim.
\newblock Cure: An efficient clustering algorithm for large databases.
\newblock In {\em Proceedings of the 1998 ACM SIGMOD International Conference
  on Management of Data}, SIGMOD '98, page 73–84, New York, NY, USA, 1998.
  Association for Computing Machinery.

\bibitem{ref2_Guha1999}
S.~Guha, R.~Rastogi, and K.~Shim.
\newblock Rock: a robust clustering algorithm for categorical attributes.
\newblock In {\em Proceedings 15th International Conference on Data Engineering
  (Cat. No.99CB36337)}, pages 512--521, 1999.

\bibitem{ref11_Hinneburg2007}
A.~Hinneburg and H.-H. Gabriel.
\newblock Denclue 2.0: Fast clustering based on kernel density estimation.
\newblock In {\em Proceedings of the 7th International Conference on
  Intelligent Data Analysis}, IDA'07, page 70–80, Berlin, Heidelberg, 2007.
  Springer-Verlag.

\bibitem{ref6_Huang2003}
Z.~Huang and M.~K. Ng.
\newblock A note on k-modes clustering.
\newblock {\em J. Classif.}, 20(2):257–261, sep 2003.

\bibitem{ref7_Ward1963}
J.~H.~W. Jr.
\newblock Hierarchical grouping to optimize an objective function.
\newblock {\em Journal of the American Statistical Association},
  58(301):236--244, 1963.

\bibitem{ref12_Karypis1999}
G.~Karypis, E.-H. Han, and V.~Kumar.
\newblock Chameleon: hierarchical clustering using dynamic modeling.
\newblock {\em Computer}, 32(8):68--75, 1999.

\bibitem{ref18_KruskalWish1978}
J.~Kruskal and M.~Wish.
\newblock {\em {Multidimensional Scaling}}.
\newblock Sage Publications, 1978.

\bibitem{ref15_Nadler2008}
B.~Nadler.
\newblock {Finite sample approximation results for principal component
  analysis: A matrix perturbation approach}.
\newblock {\em The Annals of Statistics}, 36(6):2791 -- 2817, 2008.

\bibitem{ref8_Park2009}
H.-S. Park and C.-H. Jun.
\newblock A simple and fast algorithm for k-medoids clustering.
\newblock {\em Expert Systems with Applications}, 36(2, Part 2):3336--3341,
  2009.

\bibitem{ref19_rao2018}
A.~R. Rao, S.~Garai, D.~Clarke, and S.~Dey.
\newblock A system for exploring big data: an iterative k-means searchlight for
  outlier detection on open health data.
\newblock In {\em 2018 International Joint Conference on Neural Networks
  (IJCNN)}, pages 1--8, 2018.

\bibitem{ref20_Rao2021}
A.~R. Rao, S.~Garai, S.~Dey, and H.~Peng.
\newblock Piks: A technique to identify actionable trends for policy-makers
  through open healthcare data.
\newblock {\em SN Computer Science}, 2:1--22, 2021.

\bibitem{ref16_Shen2013}
D.~Shen, H.~Shen, and J.~S. Marron.
\newblock A general framework for consistency of principal component analysis,
  2013.

\bibitem{ref13_Geer2008}
S.~A. van~de Geer.
\newblock {High-dimensional generalized linear models and the lasso}.
\newblock {\em The Annals of Statistics}, 36(2):614 -- 645, 2008.

\bibitem{ref3_Zaki1997}
M.~J. Zaki, S.~Parthasarathy, M.~Ogihara, and W.~Li.
\newblock New algorithms for fast discovery of association rules.
\newblock Technical report, USA, 1997.

\bibitem{ref9_Zhang1996}
T.~Zhang, R.~Ramakrishnan, and M.~Livny.
\newblock Birch: An efficient data clustering method for very large databases.
\newblock In {\em Proceedings of the 1996 ACM SIGMOD International Conference
  on Management of Data}, SIGMOD '96, page 103–114, New York, NY, USA, 1996.
  Association for Computing Machinery.

\end{thebibliography}
}

%








\end{document}